\documentclass[10pt,conference]{IEEEtran}

\usepackage{amsmath,amsfonts,amsthm}
\usepackage{graphicx}
\usepackage{xcolor}
\usepackage{enumitem}
\usepackage[export]{adjustbox}
\usepackage{subfig,caption}
\usepackage{url}
\usepackage{array,multirow,cellspace,booktabs}
\usepackage{icomma}
\usepackage{cite}
\usepackage{setspace}
\usepackage{array}
\usepackage{algorithm,algorithmic,mdframed}
\usepackage{amssymb}

\newtheoremstyle{mytheoremstyle}
  {\topsep} 
  {\topsep} 
  {\itshape} 
  {} 
  {\bfseries} 
  {} 
  {.5em} 
  {} 

\theoremstyle{mytheoremstyle}

\newtheorem{corollary}{Corollary}
\newtheorem{definition}{Definition}

\newcommand{\stepskip}{\medskip}

\DeclareMathOperator{\area}{area}
\DeclareMathOperator{\hull}{hull}
\DeclareMathOperator{\perimeter}{perimeter}

\usepackage{soul}

\begin{document}

\title{Hierarchy-of-Visual-Words: a Learning-based Approach for Trademark Image Retrieval}

\newif\iffinal
\finaltrue
\newcommand{\cmtid}{99999}


\iffinal


\author{\IEEEauthorblockN{V\'{i}tor N. Louren\c{c}o, Gabriela G. Silva, Leandro A. F. Fernandes}
\IEEEauthorblockA{Instituto de Computa\c{c}\~{a}o, Universidade Federal Fluminense\\
Niter\'{o}i, Rio de Janeiro, Brazil\\
\{vitorlourenco, gabrielagomessilva\}@id.uff.br, laffernandes@ic.uff.br}}


%

\else
  \author{Sibgrapi paper ID: 35 \\ }
\fi

\maketitle

\begin{abstract}
In this paper, we present the Hierarchy-of-Visual-Words~(HoVW), a novel trademark image retrieval~(TIR) method that decomposes images into simpler geometric shapes and defines a descriptor for binary trademark image representation by encoding the hierarchical arrangement of component shapes. The proposed hierarchical organization of visual data stores each component shape as a visual word. It is capable of representing the geometry of individual elements and the topology of the trademark image, making the descriptor robust against linear as well as to some level of nonlinear transformation. Experiments show that HoVW outperforms previous TIR methods on the MPEG-7 CE-1 and MPEG-7 CE-2 image databases.
\end{abstract}


\IEEEpeerreviewmaketitle

\section{Introduction}
\label{sec:introduction}

Trademark images are complex patterns consisting of graphical or figurative shape patterns~(device-mark), text words or phrases~(word-in-mark), or both. Trademark images carry not only the identification meaning but, also, the reputation and the quality meanings of the associated product or service. Thus, it is of the intrinsic interest of companies to ensure the ownership and exclusive use of their trademark images. The design of automatic trademark image retrieval~(TIR) systems has been an active research topic~\cite{Kato1990, Wu1996, Eakins1997, Wei2009, Anuar2013, Qi2010, Liu2017, Sidiropoulos2011, Yang2006, Alajlan2006, Alajlan2008} due to the complexity of manual trademark image matching analysis.

All these artificially-produced images are designed to have a visual impact and consisting of multiple elements, which may be closed regions, lines, or areas of texture. Existing TIR systems, however, typically treat trademark images as indivisible structures by computing descriptors integrating global and local image features~\cite{Kato1990, Wu1996, Eakins1997, Wei2009, Anuar2013, Qi2010} or by partitioning the image~\cite{Liu2017, Sidiropoulos2011, Yang2006} without considering the distribution of their component shapes. Such a practice has been successful in~retrieving near-duplicated images but may fail in detecting similar instances that preserve the topology of their components without conserving the relative location of their elements.

\begin{figure*}[!t]
    \centering
    \includegraphics[width=\textwidth]{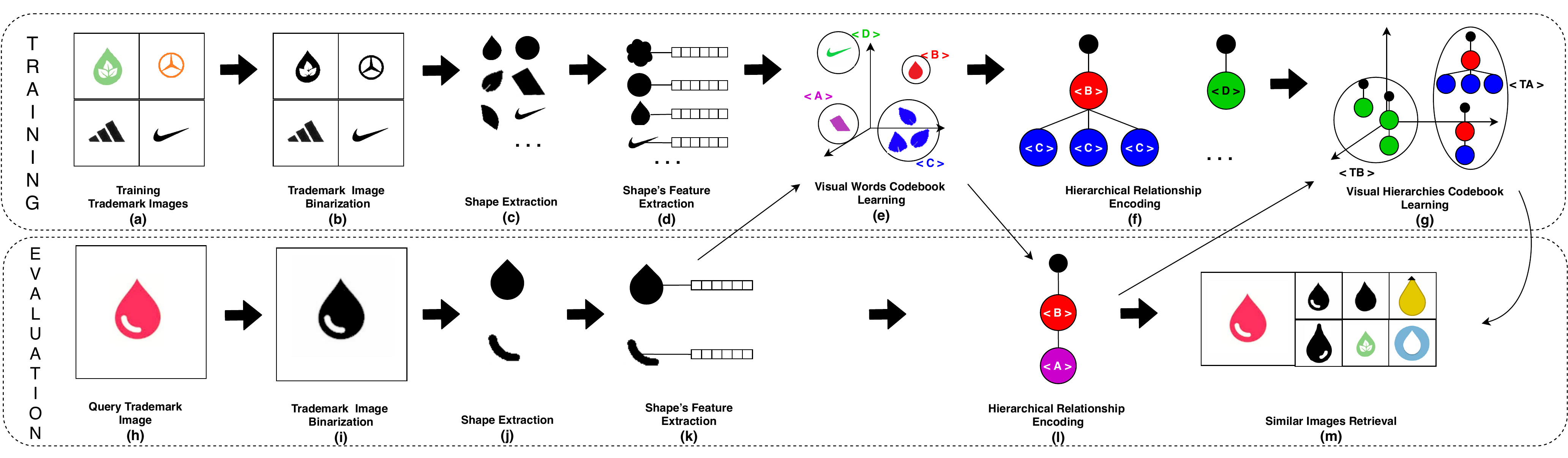}
    \caption{Overview of the training and evaluation stages of the Hierarchy-of-Visual-Words~(HoVW) approach.}
    \label{fig:hovw}
\end{figure*}

This paper proposes a novel TIR method called Hierarchy-of-Visual-Words~(HoVW). The key insights of our solution is that shape is probably the single most important feature used by human observers to characterize an image~\cite{Biederman1987}.
Also, image structure and the layout of individual image elements are essential when judging similarity~\cite{Biederman1987}.
Therefore, we have designed the HoVW approach as a method that takes component shapes, image structure, and layout of individual image elements into account while computing descriptors of trademark images. Fig.~\ref{fig:hovw} shows the main steps of the HoVW. In the training stage, our approach decomposes the set of training trademark images~(a) into simple component shapes~(b-c) and learns a codebook of visual words for those shapes~(d-e). The hierarchical arrangement of components within each image leads to the representation of trademarks as trees of visual words~(f). Then, our approach learns a codebook of visual hierarchies~(g), which defines a labeling system for trademark image representation. In the evaluation stage, the HoVW uses the visual words codebook to encode the simple shapes extracted from the query image~(h-k). Next, the hierarchical relationship of its components is enconded~(l), and the visual hierarchies codebook is used to accelerates the retrieval of similar images from the database~(m). The feature vectors in~(d) and~(k) are comprised of Zernike moments~(ZM), circularity, average bending energy, eccentricity, and convexity. The dissimilarity between two hierarchies produced in~(f) and~(l) is computed using an efficient tree edit distance algorithm~\cite{Pawlik2016}. Clustering in~(e) and~(g) are performed using, respectively, point-based $k$-means~\cite{Lloyd1982} and mean shift for distance matrices~\cite{Comaniciu2002}.


The main contributions of this paper include:
\begin{enumerate}
    \item[i)] A new learning-based framework for the hierarchical representation of elements in binary images; and
    \item[ii)] Its application on trademark image description and retrieval from image databases.
\end{enumerate}

We have performed experiments using the popular MPEG-7 CE-1 and MPEG-7 CE-2 image databases to compare the efficiency of our method against state-of-the-art solutions in TIR tasks. Precision-recall curves show that the proposed hierarchical decomposition leads to smaller dissimilarity measures between trademarks of the same class than other solutions. In other words, the query image is expected to be closer to similar images in the feature space while performing TIR for judging trademark infringement.

\section{Related Work}
\label{sec:related-work}

The first systems to implement content-based retrieval for trademark images were developed by Kato \emph{et al.}~\cite{Kato1990}, Wu \emph{et al.}~\cite{Wu1996} and Eakins \emph{et al.}~\cite{Eakins1997}. The TRADEMARK system~\cite{Kato1990} uses the spatial distribution, spatial frequency, local correlation, and local contrast of pixel blocks as visual features. The STAR system~\cite{Wu1996} represents trademarks as structural patterns consisting of word-in-mark and device-mark information. The former is manually informed by the user. The later includes Fourier descriptor of the shape, moment invariants, and grey level projections. ARTISAN~\cite{Eakins1997} applies a structural pattern similar to STAR's and incorporates principles derived from Gestalt psychology to cope with device-only marks.

Some solutions integrate global features to capture the gross essence of the shapes and local features to describe the interior details of binary trademark images. For instance, Wei \emph{et al.}~\cite{Wei2009} employ ZM to extract global features. Curvature and distance to the centroid are used as local features. Anuar \emph{et al.}~\cite{Anuar2013} also use ZM, but they employ the edge-gradient co-occurrence matrix derived from the contour information as the local descriptor. Qi \emph{et al.}~\cite{Qi2010} use the histogram of centroid distances and a region descriptor based on improved feature points matching and the spatial distribution of feature points to avoid the calculation of ZM. It is important to emphasize that, in contrast to those approaches, the final descriptor produced by HoVW does not encode local and global information separately. The multilevel organization of the data gives a global representation at some level of detail, while each node of the structure represents local information.

In recent work, Liu \emph{et al.}~\cite{Liu2017} proposed a hierarchical region feature descriptor where the binary trademark image is iteratively partitioned into progressively smaller ones along with various directions. Density, compactness, rectangularity, and eccentricity are computed for each partition. The binary image retrieval methods developed by Sidiropoulos \emph{et al.}~\cite{Sidiropoulos2011} and Yang \emph{et al.}~\cite{Yang2006} differ from Liu's \emph{et al.} work because they consider a single direction (the most descriptive one). Also, they split the image region recursively regarding the four subregions for the next partition in each iteration instead of only one. HoVW does not rely on image partitions. Our approach decomposes binary trademark images into sets of simpler component shapes and builds the hierarchical arrangement of those components. As a result, our descriptor encodes the topology of the basic image components using the hierarchy and their geometry using feature values invariant to rotation, translation, and scale.

The image retrieval system proposed by Alajlan \emph{et al.}~\cite{Alajlan2006, Alajlan2008} uses a structured representation called Curvature Tree to encode both shape and topology of objects and holes comprising a binary image. Our approach uses a different tree structure to represent topology. Also, our tree dissimilarity measure is based on tree edit distance~\cite{Pawlik2016} instead of maximum similarity subtree isomorphism~\cite{Pelillo19999}.

The use of codebooks and structured representation of images was also considered by Silva \emph{et al.}~\cite{Silva2013, Silva2017}. However, in this case, local information of grayscale images is extracted by Hessian Affine~\cite{Mikolajczyk2002} and Scale Invariant Feature Transform (SIFT)~\cite{Lowe1999} detectors. As a result, Silva's \emph{et al.} approach is not suitable for low-textured trademarks. Also, the visual-word arrangement is defined by a planar graph derived from the Delaunay triangulation of feature points instead of the tree structure applied by our approach to organizing the component shapes of the trademark image.

Recent approaches based on neural network advances are presented by Perez~\emph{et al.}~\cite{Perez2018} and Lan~\emph{et al.}~\cite{Lan2018}. Lan's~\emph{et al.} introduces the use of deep Convolutional Neural Networks~(CNN) features aided by constraint theory. Perez's~\emph{et al.} approach resides on the use of pretrained VGG19 models for learning visual and conceptual similarities. Both approaches require a set of classes and the original trademark images with modifications to train the CNN models and learn concepts of similarity. Perez~\emph{et al.}, for instance, was aided by human experts to define the set of classes on each trained model, which could introduce human bias and possibly limit the model capability to learn unknown concepts. Our approach learns synthetic classes, \emph{i.e.},~a codebook, without human interference, through the similarity between the tree-structured shapes.

\section{Hierarchy-of-Visual-Words (HoVW)}
\label{sec:hovw}

As a learning-based approach, the HoVW framework is comprised of training and evaluation stages. Fig.~\ref{fig:hovw} illustrates the main steps of each one of them.

\stepskip
\noindent
\textbf{Trademark image binarization} is performed at both stages of HoVW (Figs.~\ref{fig:hovw}~(b) and~(i)). It consists of converting digital trademark images into binary images from which component shapes will be extracted. In this step, we first convert a given color image to grayscales. Then, we apply a median filter~\cite{Huang1979} to reduce impulsive noise and a bilateral filter~\cite{Tomasi1998} to remove texture without losing overall shapes since sharp edges are preserved. The final binary image is obtained by applying Otsu's method~\cite{Otsu1979} on the textureless grayscale image.

\stepskip
\noindent
\textbf{Shape extraction} aims to split binary images into objects and holes, \emph{i.e.,}~their component shapes (Figs.~\ref{fig:hovw}~(c) and~(j)).

\begin{definition}
    \label{def:object_and_hole}
    (Object and hole): Objects and holes are connected sets of, respectively, foreground and background pixels.
\end{definition}

\begin{figure}[!t]
    \centering
    \hfill
    \includegraphics[scale=0.6, valign=t]{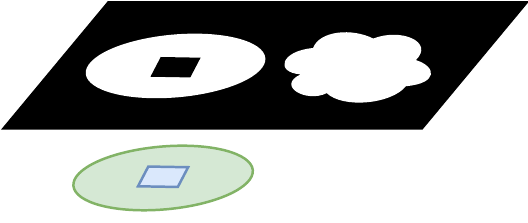}
    \includegraphics[scale=0.6, valign=t]{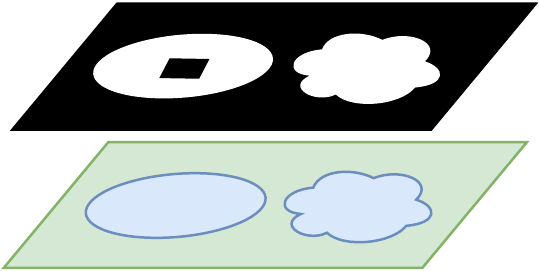}
    \hfill
    \includegraphics[scale=0.6, valign=t]{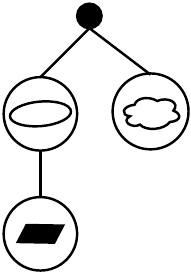}
    \hfill
    \caption{Examples of inclusion and exclusion relationships: (top, left and center)~parallel orthographic projection of a binary image; (bottom, left)~the small square hole is in included in the circular object; (bottom, center)~the circle and the cloud are included in the rectangle and exclude each other; and (right)~the hierarchical relationship of the component shapes.}
    \label{fig:hierarchical_relationship}\vspace{-12pt}
\end{figure}

For instance, in Fig.~\ref{fig:hierarchical_relationship}~(top), all the pixels that are set to white are foreground pixels, while the background pixels are set to black. Thus, this image includes two objects (the circle and the cloud) and two holes (the outer rectangle and the square inside the circle).

In this work, we extract shapes by using the border-following method proposed by Suzuki and Abe~\cite{Suzuki1985}. Their approach applies a border labeling mechanism capable of describing the relationship among the outer borders and the hole borders, capturing the topological structure of a given binary image.

Fig.~\ref{fig:suzuki_and_abe_algorithm} presents the algorithm used to extract the component shapes of a given trademark image. The algorithm, first, associates the input binary image $\mathcal{I}$ with a decomposing list $M$ (line 2). After, the main iterative process (lines 3 to 10) segments each element $m_{i}$ on the decomposing list $M$ concerning their foreground and background pixels (lines 5 and 6). From the background, the procedure extracts the holes' shapes and associate them with the component shapes' list (lines 7 and 8). The foreground shapes are used in the next iterations (lines 5 and 9) until all component shapes have been extracted from the initial binary trademark image.

\stepskip
\noindent
\textbf{Shape's feature extraction} consists of building a feature vector for each component shape of a given trademark image (Figs.~\ref{fig:hovw}~(d) and~(k)). These $29$-dimension feature vectors combine region-based and contour-based descriptors.

Shape's region is described by the $25$~moments of the Zernike polynomials~(ZM) of order $p$ from $0$ to $8$:
\begin{equation}
    \label{eq:Zernike_moment}
    Z_{p,q} = \frac{p + 1}{\pi} \sum_{\rho}{\sum_{\theta}{V_{p,q}\left(\rho, \theta\right)^{*} \, \mathcal{I}\left(\rho, \theta\right)}} \text{,}
\end{equation}
where \mbox{$\rho = \sqrt{x^2 + y^2}$} is the length of vector from origin to pixel \mbox{$(x, y)$}, $\theta$ is the angle between the vector defining $\rho$ and the $x$-axis in the counter clockwise direction and $V_{p,q}\left(\rho, \theta\right)$ is a Zernike polynomial of order $p$ with repetition $q$ that forms a complete set over the interior of the unit disk inscribing the component shape:
\begin{equation}
    \label{eq:Zernike_polynomial}
    V_{p,q}\left(\rho, \theta\right) = R_{p,q}\left(\rho\right) \exp\left(-\mathrm{i} \, q \, \theta\right) \text{.}
\end{equation}
In Equation~(\ref{eq:Zernike_polynomial}), \mbox{$\mathrm{i} =\sqrt{-1}$}, $p$ is a non-negative integer value and $q$ is a positive integer subject to \mbox{$p - \left|q\right| = \text{even}$} and \mbox{$q \leq p$}. The radial polynomial $R_{p,q}\left(\rho\right)$ is defined as:
\begin{equation}
    \label{eq:radial_polynomial}
    R_{p,q}\left(\rho\right) = \sum_{s=0}^{(p - \left|q\right|)/2}\frac{\left(-1\right) \left(p - s\right)!}{s! \left(p + \left|q\right| - s\right)! \left(p - \left|q\right| - s\right)} \rho^{p - 2 s} \text{.}
\end{equation}

ZM are rotation invariant by construction. We use Khotanzad and Hong's approach~\cite{Khotanzad1990} to obtain invariance to translation and scale, too.

We have used four measures to describe the shape's contour~\cite{Mingqiang2008}:
\begin{itemize}[noitemsep, nosep, leftmargin=*]
    \item \emph{Circularity} defines the relation between the perimeter of the shape $\mathcal{S}$ and its area:
    \begin{equation}
        \label{eq:circularity}
        d_{c} = \frac{\perimeter\left(\mathcal{S}\right)^2}{4 \pi \area\left(\mathcal{S}\right)} \text{;}
    \end{equation}
    \item \emph{Average bending energy} defines the mean sum of the shape's curvature:
    \begin{equation}
        \label{eq:average_bending_energy}
        d_{b} = \frac{1}{N} \sum_{t = 0}^{N-1} K(t)^2 \text{,}
    \end{equation}
    where $K(t)$ is the curvature function, $t$ is the arc length parameter and $N$ is the number of points on the contour~\cite{Young1974};
    \item \emph{Eccentricity} characterizes the statistical distribution of contour points around the principal axes of shape's contour:
    \begin{equation}
        \label{eq:eccentricity}
        d_{e} = \frac{\lambda_2}{\lambda_1} \text{,}
    \end{equation}
    where \mbox{$\lambda_1$} and \mbox{$\lambda_2$} are the eigenvalues of the covariance matrix of the set of contour points of $\mathcal{S}$, for \mbox{$\lambda_1 > \lambda_2$};
    \item \emph{Convexity} defines the relation between the perimeter of the convex hull and the perimeter of the shape:
    \begin{equation}
        \label{eq:convexity}
        d_{x} = \frac{\hull\left(\mathcal{S}\right)}{\perimeter\left(\mathcal{S}\right)} \text{.}
    \end{equation}
\end{itemize}

\begin{figure}[!b]
    \centering
    \small
    \begin{mdframed}
        \begin{algorithmic}[1]
            \REQUIRE{Binary trademark image $\mathcal{I}$, where the background pixels are black, \emph{i.e.}, $0$ bit, and the foreground pixels are white, \emph{i.e.} $1$ bit.}
            \ENSURE{Component shapes of the trademark image.}
            \STATE{$S \gets \{\}$}
            \STATE{$M \gets \{\mathcal{I}\}$}
            \WHILE{$M \neq \varnothing$}
                \STATE{$m_{i} \gets$ The first element in $M$}
                \STATE{$F \gets$ Foreground pixels in $m_{i}$}
                \STATE{$B \gets$ Background pixels in $m_{i}$}
                \STATE{$H \gets$ Hole shapes extracted from $B$}
                \STATE{$S \gets S \cup H$}
                    \STATE{$M \gets \left(M \setminus \{m_{i}\}\right) \cup F$}
            \ENDWHILE
            \RETURN{$S$}
        \end{algorithmic}
    \end{mdframed}
    \caption{Algorithm for extracting the components shapes of a given trademark image.}
    \label{fig:suzuki_and_abe_algorithm}
\end{figure}

We have chosen the descriptors mentioned above because they showed to be robust and have low computational cost. Also, they are invariant to rotation, translation, and scale.

\stepskip
\noindent
\textbf{Visual words codebook learning} is performed only during the training stage of the HoVW approach (Fig.~\ref{fig:hovw}~(e)). In this step, we apply $k$-means clustering~\cite{Lloyd1982} on the feature vectors computed for the component shapes of the training images. The resulting clusters are a general representation of the shapes, in which each cluster acts as a word in the codebook~$\Lambda$ that assigns a given shape to a learned visual word by using the Euclidean distance between the shape's feature vector and the cluster's centroid.

The use of $k$-means or similar clustering techniques for defining codebooks of visual words is not new. The original contribution of our approach is the way the simple component shapes of the binary image is arranged hierarchically, combining topological and invariant geometrical information in the same representation.

\stepskip
\noindent
\textbf{Hierarchical relationship encoding} is a key step of the HoVW framework (Fig.~\ref{fig:hovw}~(f) and~(l)). For each binary image, this step produces a tree structure induced by:

\begin{definition}
    \label{def:shape_inclusion}
    (Shape inclusion): A shape $\mathcal{A}$ is said to be included in a shape $\mathcal{B}$ if and only if $\mathcal{A}$ is a hole surrounded by object $\mathcal{B}$ or $\mathcal{A}$ is an object surrounded by hole $\mathcal{B}$.
\end{definition}

\begin{definition}
    \label{def:shape_exclusion}
    (Shape exclusion): Shapes $\mathcal{A}$ and $\mathcal{B}$ exclude each other if and only if they
    are included in shape $\mathcal{C}$.
\end{definition}

\begin{corollary}
    \label{cor:visual_hierarchy}
    (Visual hierarchy): The recursive inclusion and exclusion relationship of objects and holes yields the hierarchical organization of visual data into a tree structure where each node corresponds to one shape, and it is related to its ancestor node by inclusion and to its siblings by exclusion.
\end{corollary}

For instance, in Fig.~\ref{fig:hierarchical_relationship}~(left), the small square hole is included in the circular object, while in Fig.~\ref{fig:hierarchical_relationship}~(center) the circle and cloud exclude each other but are included in the black rectangle. According to Corollary~\ref{cor:visual_hierarchy}, the hierarchical relationship of those shapes leads to the tree in Fig.~\ref{fig:hierarchical_relationship}~(right).

In the HoVW framework, each node of the visual hierarchy stores a reference to the word that better describes the respective node's shape in the codebook of visual words. This way, the hierarchy encodes topology while nodes encode the geometry of component shapes.

\stepskip
\noindent
\textbf{Visual hierarchies codebook learning} is the last step of the training stage (Fig.~\ref{fig:hovw}~(g)). It aims to discover the most suitable set of labels to represent similar visual hierarchies within a database. Those labels are used to accelerate TIR queries during the evaluation stage of the HoVW framework~(Fig.~\ref{fig:hovw}~(m)).

In this step, we compute the dissimilarity matrix of visual hierarchies representing the training images and use this matrix as the input of the mean shift clustering procedure~\cite{Comaniciu2002} with RBF kernel. Ideally, the dissimilarity measure between two visual hierarchies has to be robust against changes on the compared trademark images. Those changes include linear and non-linear transformations and the addition and removal of elements. Also, it must be computationally and memory-efficient. We satisfy those requirements by using the All Path Tree Edit Distance~(AP-TED) algorithm developed by Pawlik and Augsten~\cite{Pawlik2016}, an approach that counts the minimal-cost sequence of editing operations needed to transform a tree into another while keeping low computational cost and memory footprint.

We have modeled the costs of the \emph{rename}, \emph{insert} and \emph{remove} editing operations performed by AP-TED as follows:

\smallskip
\noindent
\emph{Rename} a node is similar to change the visual word stored by it. Thus, the cost of this operation is the Euclidean distance between the centroids of the clusters corresponding to the actual and the desired words in the codebook~$\Lambda$:
\begin{equation}
    \label{eq:relabel}
    \delta_{r}(n_a, n_b) = \mathrm{dist}_{E}(\lambda_a, \lambda_b) \text{,}
\end{equation}
where $\mathrm{dist}_{E}$ denotes the Euclidean distance, \mbox{$n_a$} and \mbox{$n_b$} are tree nodes, and \mbox{$\lambda_a$} and \mbox{$\lambda_b$} are their respective visual words.

\smallskip
\noindent
\emph{Insert} and \emph{remove} costs are calculated as functions of the mean distance between all pairs of visual words in $\Lambda$, modulated by a factor~$\alpha$. This factor is proportional to the most influential value between the deep $\mathcal{D}$ of the node $n$ in the tree and the number of siblings of $n$ at same level $\mathcal{L}$ of the tree:
\begin{equation}
    \label{eq:insert-remove}
    \delta_{x}(n) = \alpha \, \frac{2}{m \left(m - 1\right)} \sum_{i=1}^{m}\sum_{j=i+1}^{m} \mathrm{dist}_{E}(\lambda_i, \lambda_j) \text{,}
\end{equation}
where \mbox{$\alpha = \min\{\log_2^{-1}\mathcal{L}, \log_2^{-1}\mathcal{D}\}$}. In Equation~(\ref{eq:insert-remove}), $m$ is the number of visual words in $\Lambda$ and $\lambda_i$ and $\lambda_j$ are visual words in $\Lambda$.

\begin{figure}[!t]
    \centering
    \includegraphics[width=\columnwidth]{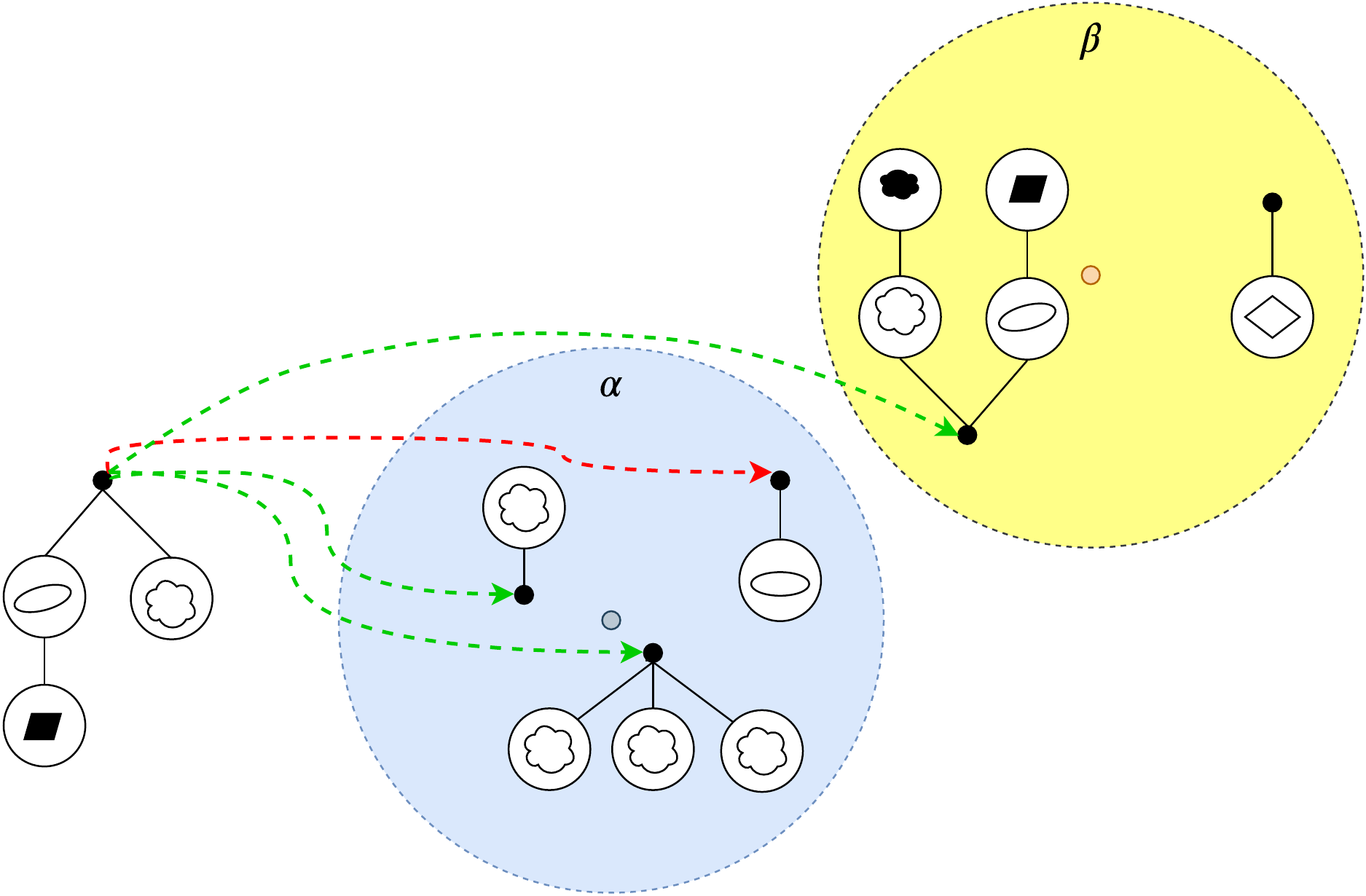}
    \caption{Example of our similar image search strategy. The query hierarchy one the left was related to hierarchies included in clusters $\alpha$ and $\beta$.}
    \label{fig:similar-search}
\end{figure}

Both cost heuristics were chosen given the spatial characteristics of the shapes' representation through visual words and the hierarchical semantics encoded in the hierarchies.

The cost of rename a node $n_a$ to a node $n_b$ is straight-forward from Equation~(\ref{eq:relabel}) as the Euclidean distances between their visual words. The heuristic of the cost of a node $n$ insertion or removal, \emph{i.e.},~$\delta_{x}(n)$, was defined based on the characteristics of the tree-structured trademark images' shape. These characteristics ensure that the deeper a shape on the hierarchy, the smaller its size compared to the whole trademark image. Also, it assures the more shapes on a tree level, the greater the amount of information on it. Then, in both cases, the relevance of a single shape is ensured to be lower. The semantics of the presented characteristics are embedded on factor $\alpha$, while the other part from Equation~(\ref{eq:insert-remove}) performs the average distance between all visual words.

\stepskip
\noindent
\textbf{Similar images retrieval} step receives the hierarchical representation of the query image
as input (Fig.~\ref{fig:hovw}~(m)) and returns a set of related images. Recall that the proposed representation is a tree structure. As a result, conventional point-based searching strategy for retrieving the $k$-nearest neighbors in feature space cannot be used with our approach. We overcome this issue by using the codebook $\Theta$ of visual hierarchies and a $k$-d tree to speed up database search. Our searching strategy first compares the query image representation with the set of database entries having the same label in $\Theta$. It only looks within other sets of entries having close labels if the user wants to retrieve more images. Retrieved images are naturally presented in ascending order of dissimilarity.

An example of the used similar image search strategy is illustrated in Fig.~\ref{fig:similar-search}. In this example, the hierarchical representation of the query image (right) is associated with the cluster having label $\alpha$. To retrieve all similar images of the query image, first, all images within the same $\alpha$~cluster are retrieved according to the measured dissimilarity. Since not all images related to the query image are within label~$\alpha$, the search strategy looks for other similar images in the cluster closest~$\alpha$, \emph{i.e.,}~cluster~$\beta$. On $\beta$~cluster, the similar missing image is found. Then, it joins the retrieved set, again, taking into account the dissimilarity between the query image and the image found.

\begin{figure}[!t]
    \centering
    \begingroup
    \setlength{\tabcolsep}{1pt}
    \renewcommand{\arraystretch}{0.5}
    \fboxsep=0mm
    \fboxrule=1pt
    \begin{tabular}{cccccc}
        \multirow{4}{*}[-10pt]{\fcolorbox{blue}{white}{\includegraphics[width=20mm]{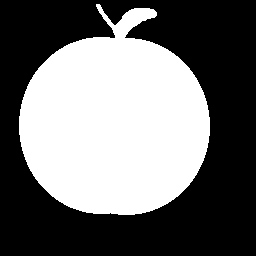}}}
        & \fcolorbox{green}{white}{\includegraphics[width=12mm]{fig/match-apple/1.png}}
        & \fcolorbox{green}{white}{\includegraphics[width=12mm]{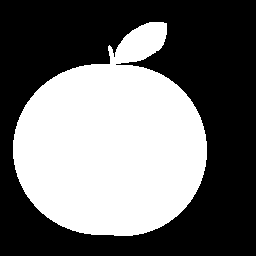}}
        & \fcolorbox{green}{white}{\includegraphics[width=12mm]{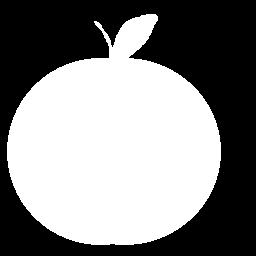}}
        & \fcolorbox{green}{white}{\includegraphics[width=12mm]{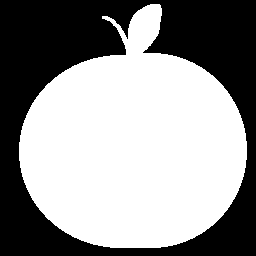}}
        & \fcolorbox{green}{white}{\includegraphics[width=12mm]{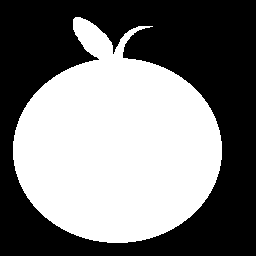}}
        \\
        & \fcolorbox{green}{white}{\includegraphics[width=12mm]{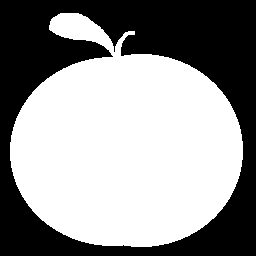}}
        & \fcolorbox{green}{white}{\includegraphics[width=12mm]{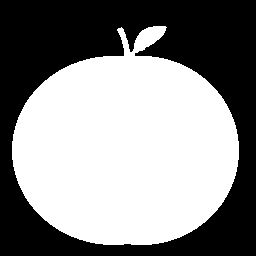}}
        & \fcolorbox{green}{white}{\includegraphics[width=12mm]{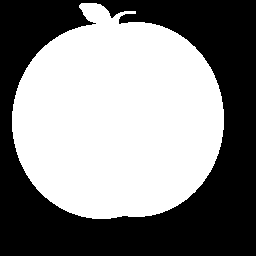}}
        & \fcolorbox{green}{white}{\includegraphics[width=12mm]{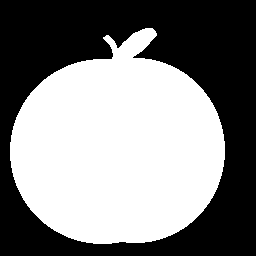}}
        & \fcolorbox{green}{white}{\includegraphics[width=12mm]{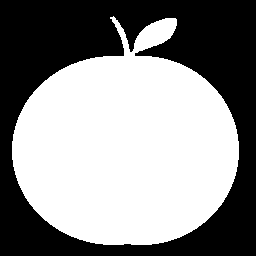}}
        \\
        & \fcolorbox{green}{white}{\includegraphics[width=12mm]{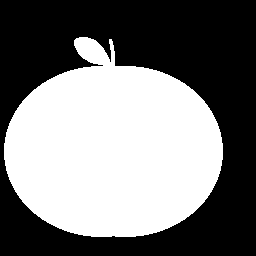}}
        & \fcolorbox{green}{white}{\includegraphics[width=12mm]{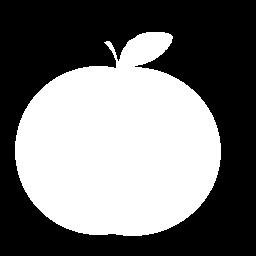}}
        & \fcolorbox{green}{white}{\includegraphics[width=12mm]{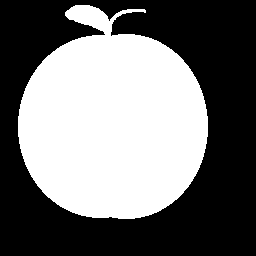}}
        & \fcolorbox{green}{white}{\includegraphics[width=12mm]{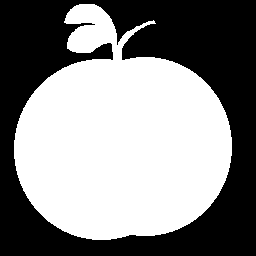}}
        & \fcolorbox{green}{white}{\includegraphics[width=12mm]{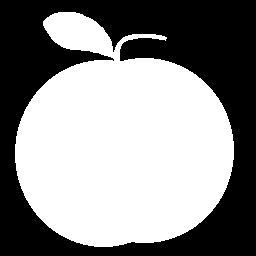}}
        \\
        & \fcolorbox{green}{white}{\includegraphics[width=12mm]{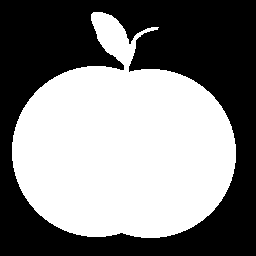}}
        & \fcolorbox{green}{white}{\includegraphics[width=12mm]{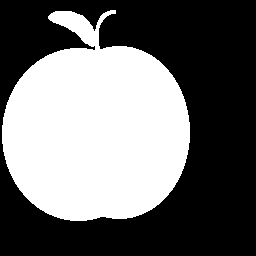}}
        & \fcolorbox{green}{white}{\includegraphics[width=12mm]{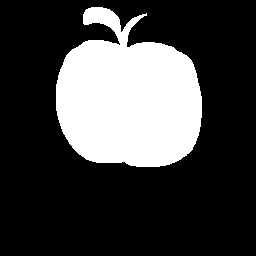}}
        & \fcolorbox{green}{white}{\includegraphics[width=12mm]{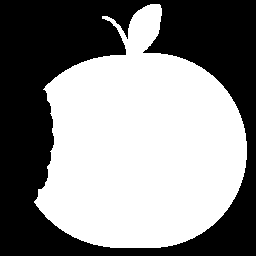}}
        & \fcolorbox{green}{white}{\includegraphics[width=12mm]{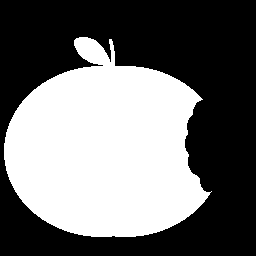}}
        \\[1em]
        \multirow{4}{*}[-10pt]{\fcolorbox{blue}{white}{\includegraphics[width=20mm]{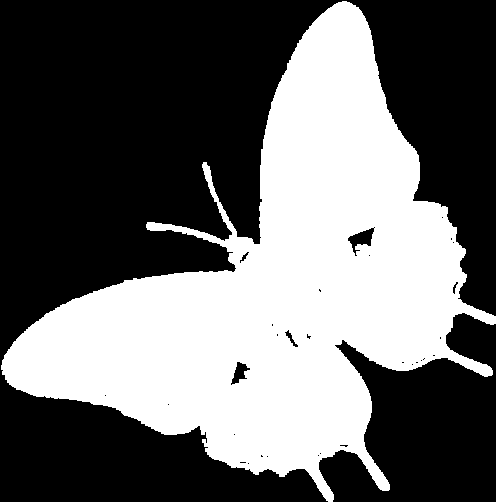}}}
        & \fcolorbox{green}{white}{\includegraphics[width=12mm]{fig/match-butterfly/1.png}}
        & \fcolorbox{green}{white}{\includegraphics[width=12mm]{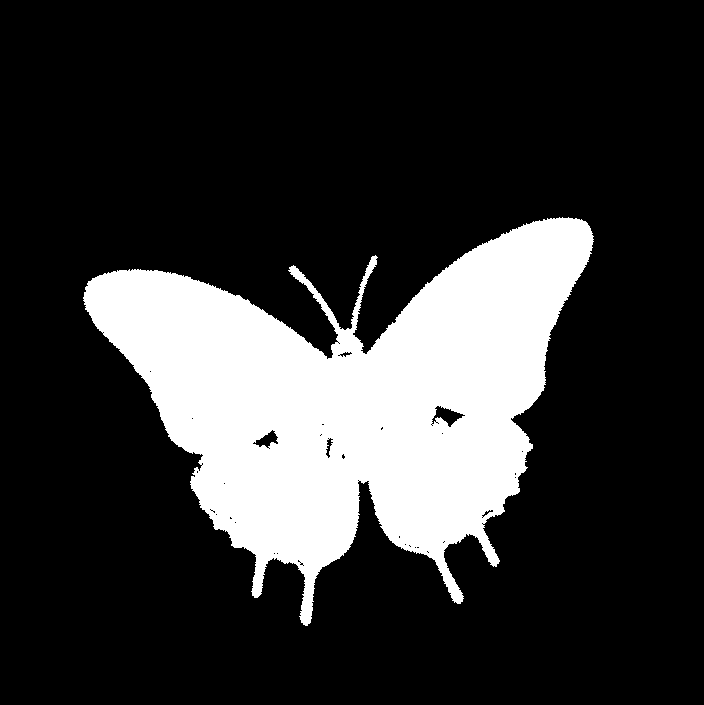}}
        & \fcolorbox{green}{white}{\includegraphics[width=12mm]{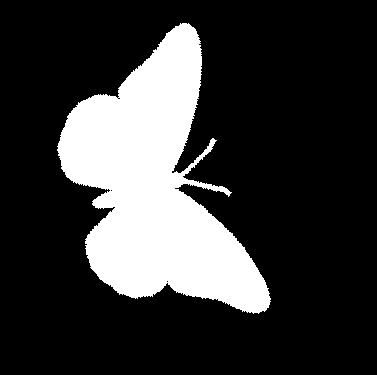}}
        & \fcolorbox{green}{white}{\includegraphics[width=12mm]{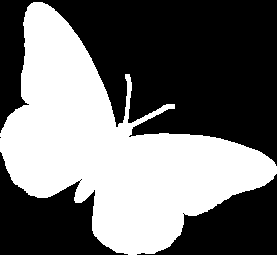}}
        & \fcolorbox{green}{white}{\includegraphics[width=12mm]{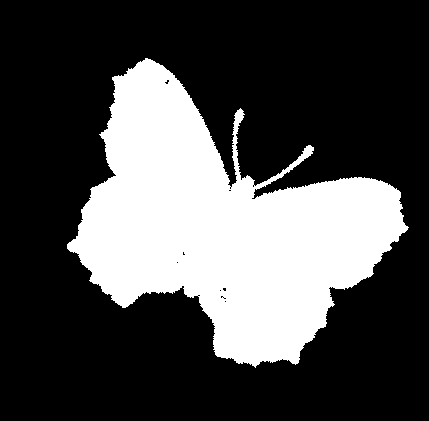}}
        \\
        & \fcolorbox{green}{white}{\includegraphics[width=12mm]{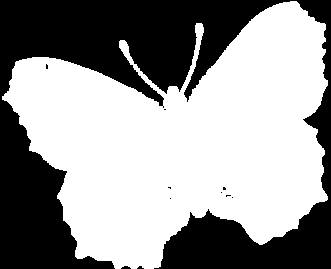}}
        & \fcolorbox{green}{white}{\includegraphics[width=12mm]{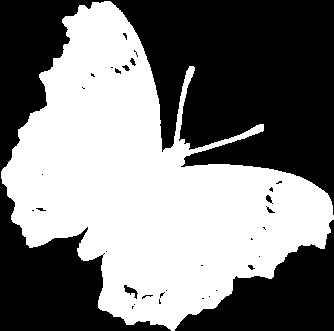}}
        & \fcolorbox{green}{white}{\includegraphics[width=12mm]{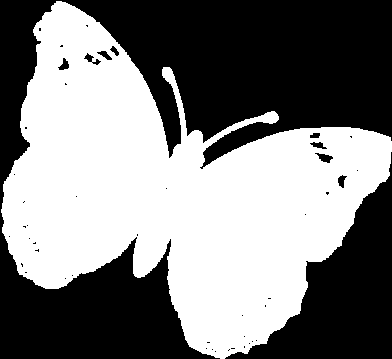}}
        & \fcolorbox{green}{white}{\includegraphics[width=12mm]{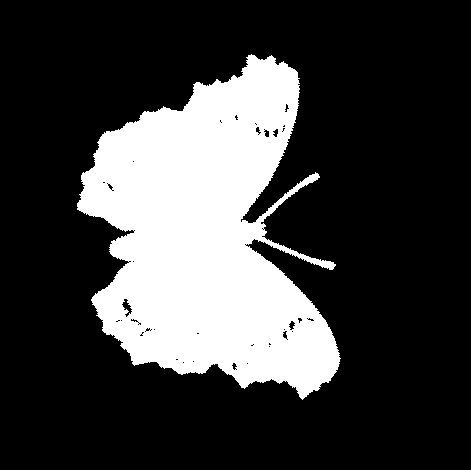}}
        & \fcolorbox{green}{white}{\includegraphics[width=12mm]{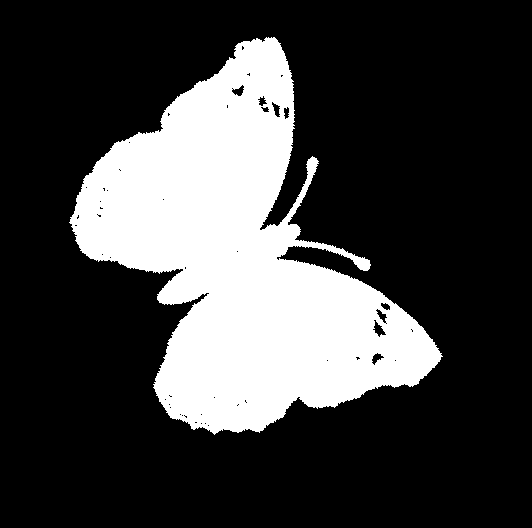}}
        \\
        & \fcolorbox{green}{white}{\includegraphics[width=12mm]{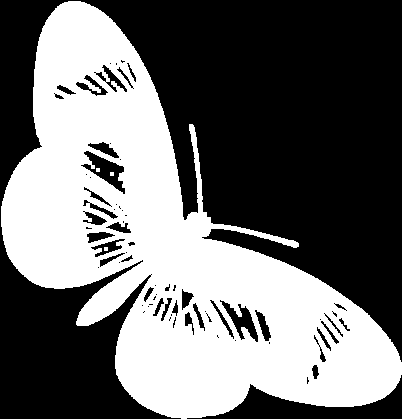}}
        & \fcolorbox{green}{white}{\includegraphics[width=12mm]{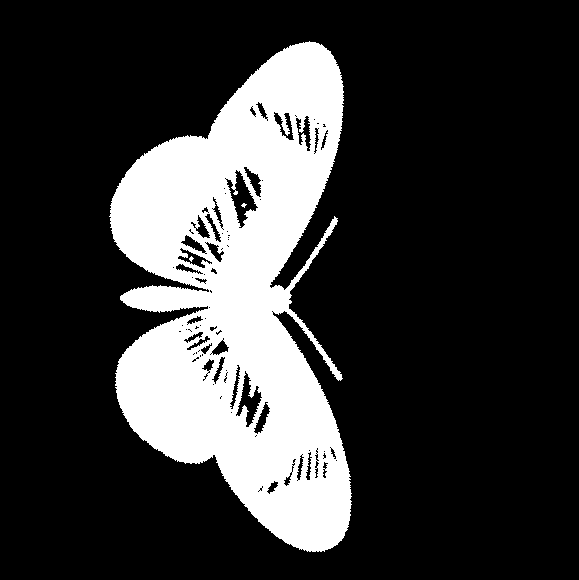}}
        & \fcolorbox{green}{white}{\includegraphics[width=12mm]{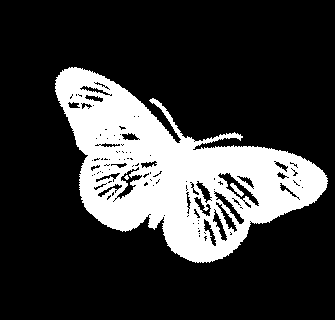}}
        & \fcolorbox{green}{white}{\includegraphics[width=12mm]{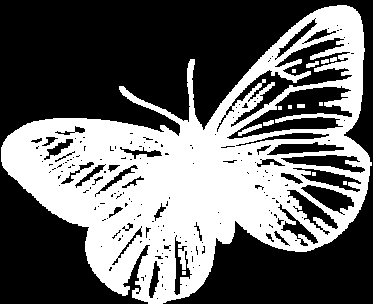}}
        & \fcolorbox{red}{white}{\includegraphics[width=12mm]{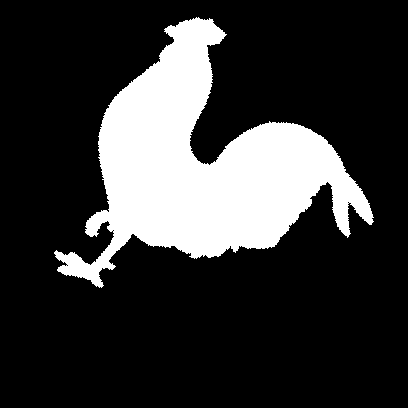}}
        \\
        & \fcolorbox{red}{white}{\includegraphics[width=12mm]{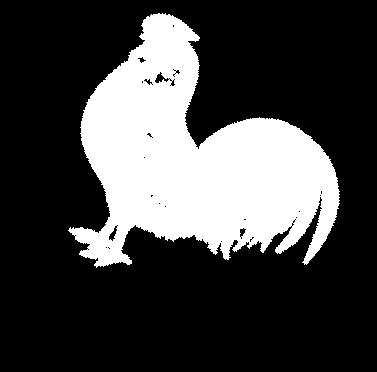}}
        & \fcolorbox{red}{white}{\includegraphics[width=12mm]{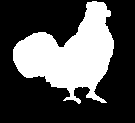}}
        & \fcolorbox{red}{white}{\includegraphics[width=12mm]{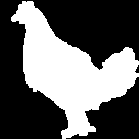}}
        & \fcolorbox{red}{white}{\includegraphics[width=12mm]{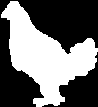}}
        & \fcolorbox{red}{white}{\includegraphics[width=12mm]{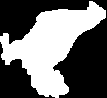}}
    \end{tabular}
    \endgroup
    \caption{The top-20 retrieval set of the HoVW for \normalsize{\texttt{apple-1.png}} and \normalsize{\texttt{butterfly-1.png}} images of the MPEG-7 CE-1 database. Green denotes images of the same class as the query entry~(blue). Red denotes different classes.}
    \label{fig:retrieval}
\end{figure}

Fig.~\ref{fig:retrieval} presents two examples of TIR performed by HoVW in the MPEG-7 CE-1 database. In the first example, the first $20$ images retrieved are of the same class as the query image. In the second examples, the $14$ first images retrieved are butterfly images. This database includes $20$ images of each class.

\section{Experiments and Results}
\label{sec:experiments-and-results}

Sections~\ref{sec:databases} and~\ref{sec:training} present the materials and methods of our experiments. In Section~\ref{sec:precision-recall_analysis}, we compare the performance of our approach to the traditional use of ZM~\cite{Khotanzad1990} and to the state-of-the-art TIR techniques proposed by Liu \emph{et al.}~\cite{Liu2017} and Anuar \emph{et al.}~\cite{Anuar2013}. In this analysis, we use precision-recall curves~\cite{Liu2009} to observe the quality of retrieval simulating the gradual detection of related samples in the database. In Section~\ref{sec:f1score_analysis}, we analyze the harmonic mean between precision and recall, \emph{i.e.,}~the $\textrm{F}_1$~score, of the aforementioned techniques and a CNN-based approach inspired by Perez's~\emph{et al.}~\cite{Perez2018} work and implemented by us. It is important to comment that, to the best of our knowledge, there is no available implementation of Perez's~\emph{et al.} technique. Also, the CNN-based approach was not included in the analysis of precision-recall curves because its outcome does not have the same semantic distance than the distance functions used by other approaches, which allows the simulation of gradual detection of related samples. Section~\ref{sec:limitations} discusses the limitations of our approach.

\subsection{Databases}
\label{sec:databases}

\begin{figure}[!b]
    \centering
    \begingroup
    \setlength{\tabcolsep}{1pt}
    \renewcommand{\arraystretch}{0.5}
    \begin{tabular}{|c|ccccccc|}
        \hline
        \multirow{2}{*}[-4pt]{\footnotesize{~~MPEG-7 CE-1~~}}
        & \includegraphics[width=8mm]{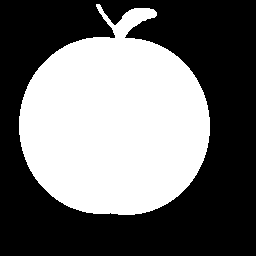}
        & \includegraphics[width=8mm]{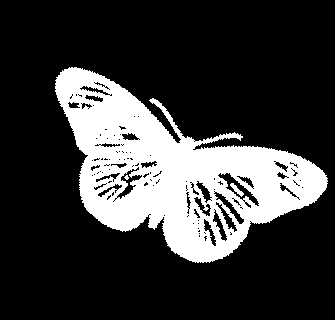}
        & \includegraphics[width=8mm]{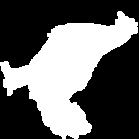}
        & \includegraphics[width=8mm]{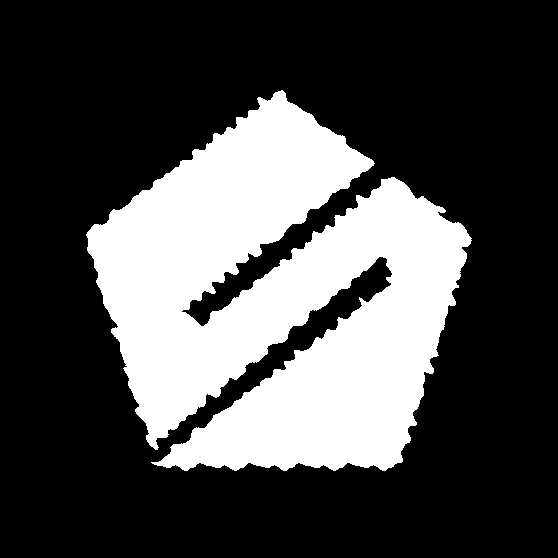}
        & \includegraphics[width=8mm]{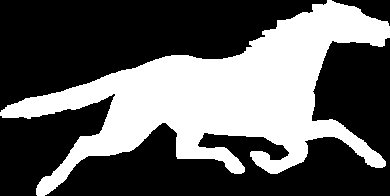}
        & \includegraphics[width=8mm]{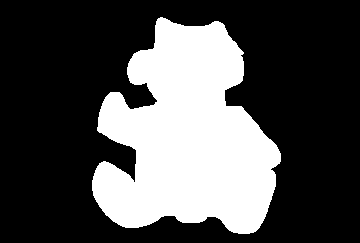}
        & \includegraphics[width=8mm]{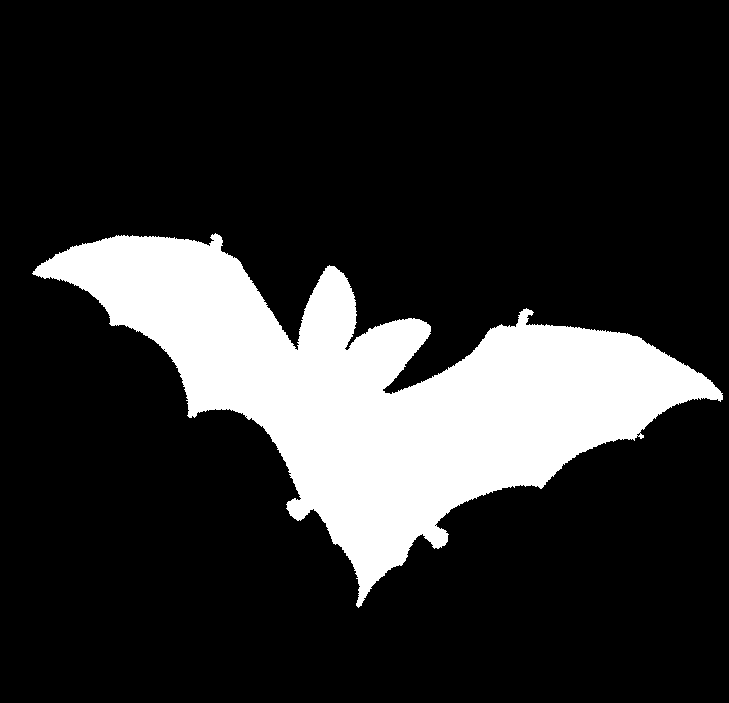}
        \\
        & \includegraphics[width=8mm]{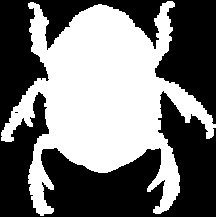}
        & \includegraphics[width=8mm]{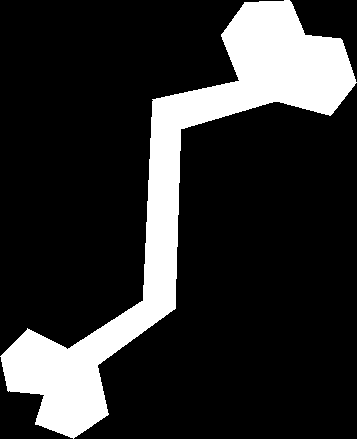}
        & \includegraphics[width=8mm]{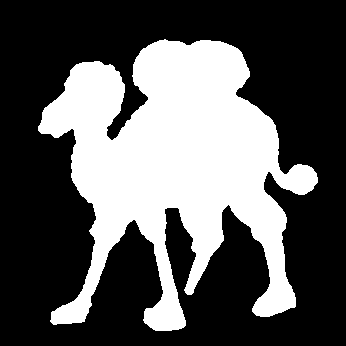}
        & \includegraphics[width=8mm]{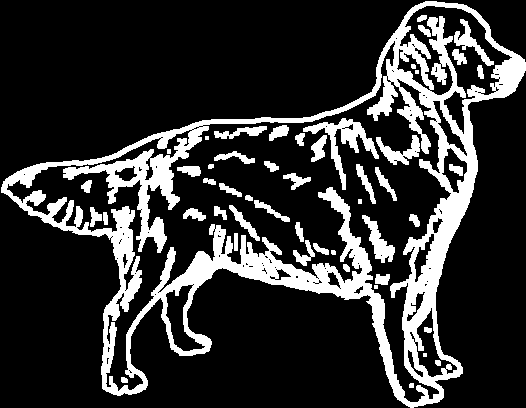}
        & \includegraphics[width=8mm]{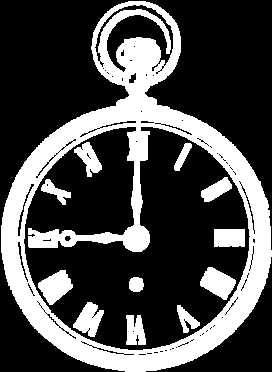}
        & \includegraphics[width=8mm]{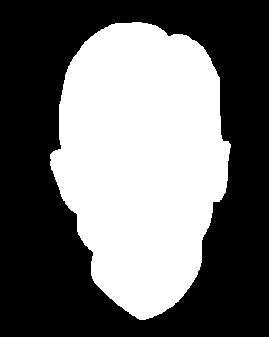}
        & \includegraphics[width=8mm]{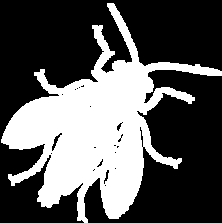}
        \\
        \hline
        \multirow{2}{*}{\footnotesize{~~MPEG-7 CE-2~~}}
        & \includegraphics[width=8mm]{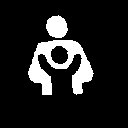}
        & \includegraphics[width=8mm]{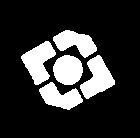}
        & \includegraphics[width=8mm]{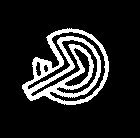}
        & \includegraphics[width=8mm]{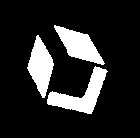}
        & \includegraphics[width=8mm]{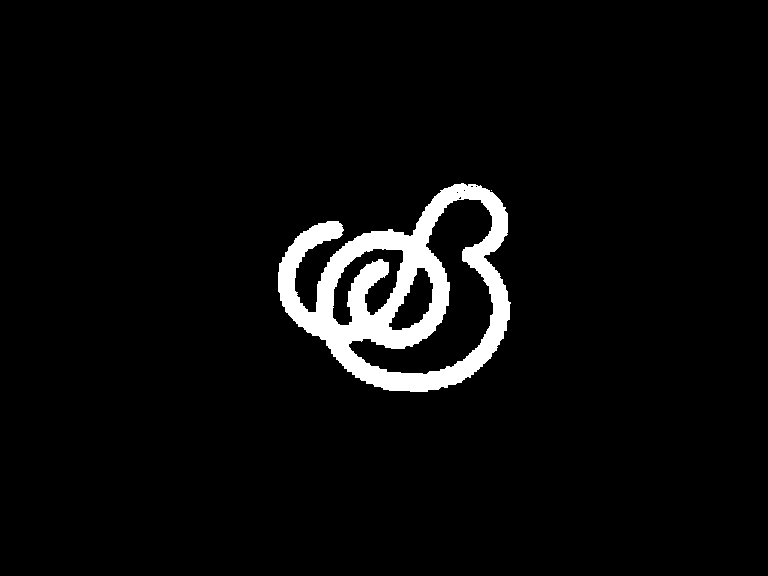}
        & \includegraphics[width=8mm]{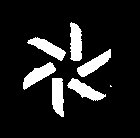}
        & \includegraphics[width=8mm]{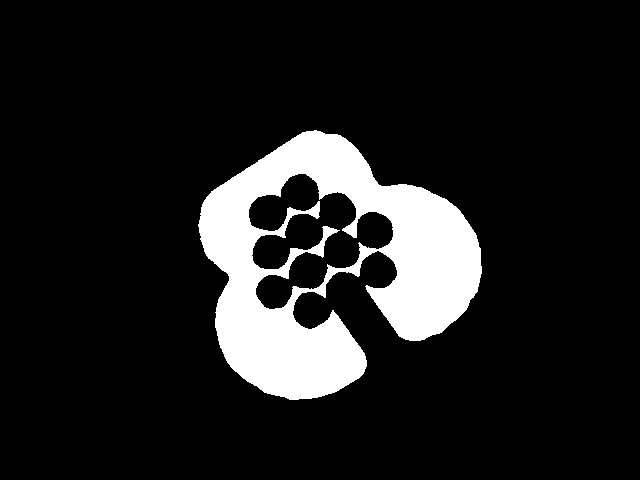}
        \\
        & \includegraphics[width=8mm]{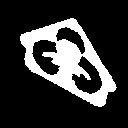}
        & \includegraphics[width=8mm]{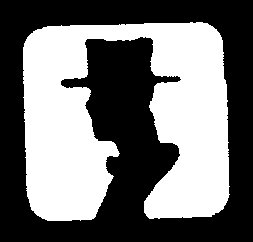}
        & \includegraphics[width=8mm]{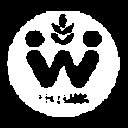}
        & \includegraphics[width=8mm]{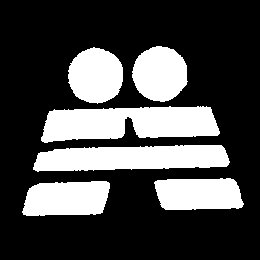}
        & \includegraphics[width=8mm]{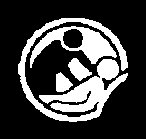}
        & \includegraphics[width=8mm]{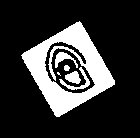}
        & \includegraphics[width=8mm]{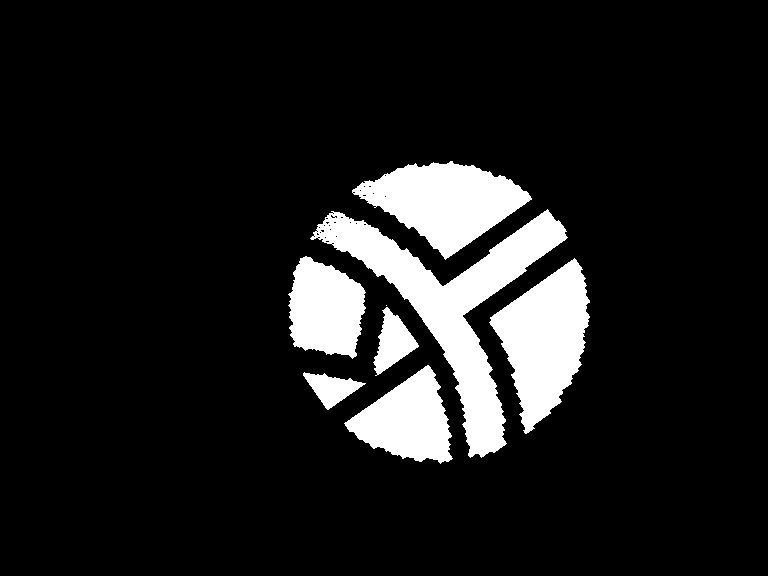}
        \\
        \hline
    \end{tabular}
    \endgroup
    \caption{Samples from the databases used in our experiments.}
    \label{fig:databases}
\end{figure}

    We have used two image databases in our experiments: \emph{MPEG-7 Core Experiment CE-Shape-1} and \emph{MPEG-7 Region Shape Dataset CE-2}~\cite{Kim2000}. The MPEG-7 CE-1 database is comprised of $1,400$ binary images organized into $70$ classes having $20$ similar images each. The MPEG-7 CE-2 database includes $871$ binary images organized into $51$ classes having from $11$ to~$21$ images each and $2,750$ images that do not belong to any category. Following Anuar~\emph{et al.}~\cite{Anuar2013} and Liu~\emph{et al.}~\cite{Liu2017}, we have used only the categorized images in our experiments because uncategorized images do not follow an overall shape pattern. Fig.~\ref{fig:databases} shows some samples from both databases.

\subsection{Implementation, Training and Evaluation}
\label{sec:training}

\stepskip
\noindent
\textbf{Liu \emph{et al.}, Anuar \emph{et al.}, and ZM Data Acquisition.} We have used the results reported by the authors in their original papers~\cite{Liu2017, Anuar2013, Wei2009} since those approaches do not require a training phase. 

\stepskip
\noindent
\textbf{HoVW Implementation and Parameterization.} We have implemented our algorithms using Python. Our~system\footnote{HoVW: \url{https://github.com/Prograf-UFF/HoVW}} performs codebook learning and efficient closest visual word searching using the implementations of $k$-means, mean shift and $k$-d tree provided by the \texttt{scikit-learn}\footnote{scikit-learn: \url{https://scikit-learn.org} library}.
Image filtering and shape extraction are performed using \texttt{OpenCV}\footnote{OpenCV: \url{https://opencv.org}}.
The dissimilarity of hierarchies is computed using the AP-TED's implementation provided by the authors\footnote{AP-TED: \url{http://tree-edit-distance.dbresearch.uni-salzburg.at}}.

We have performed median filtering using a \mbox{$5 \times 5$} window during the first step of the HoVW procedure. We skipped the application of bilateral filtering since the databases have only binary images.

\begin{figure}[!t]
    \centering
    \subfloat[][MPEG-7 CE-1, $k$ vs. MAP]{\includegraphics[width=0.8\columnwidth]{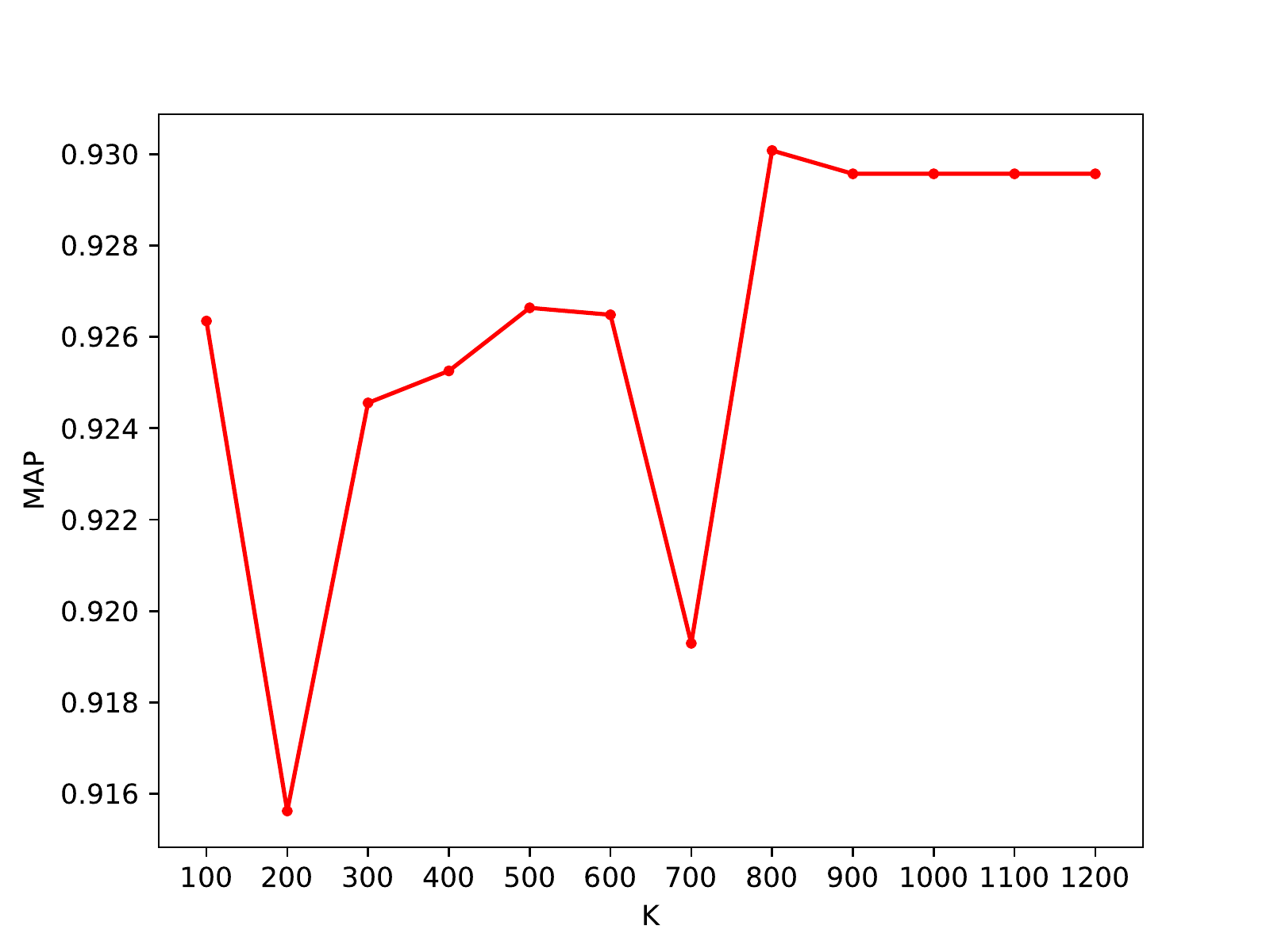}}\\%
    \subfloat[][MPEG-7 CE-1, $h$ vs. MAP]{\includegraphics[width=0.8\columnwidth]{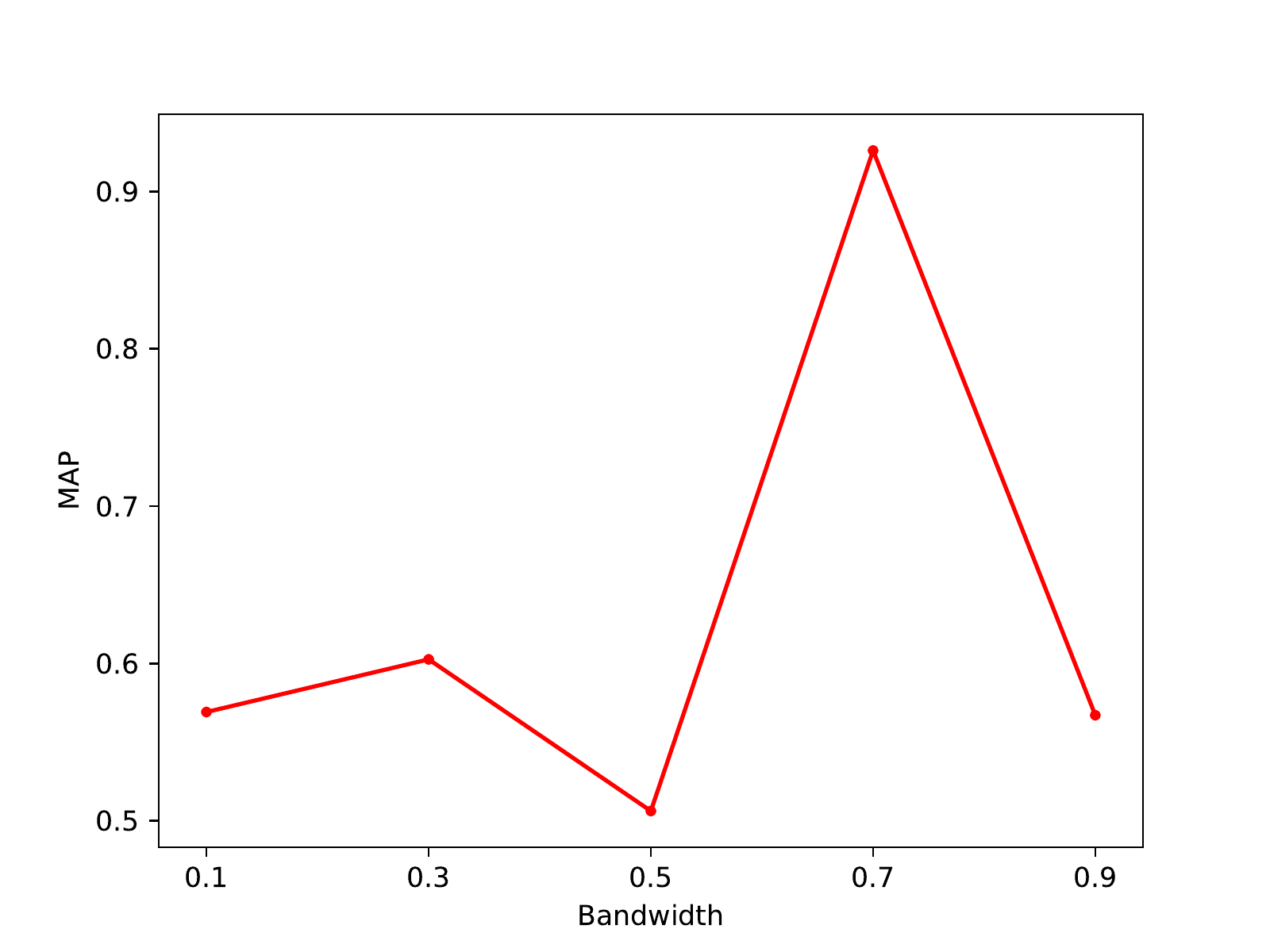}}%
    \caption{MAP values on the MPEG-7 CE-1 function of different sizes $k$ of visual words codebook~(a) and different bandwidth values $h$ for visual hierarchies codebook learning~(b).}
    \label{fig:size_and_bandwidth}
\end{figure}

The size of the visual words codebook was set to \mbox{$k = 800$} for MPEG-7 CE-1 and to \mbox{$k = 600$} for MPEG-7 CE-2 after looking for the maximum among the mean average precision~(MAP) metric values~\cite{Liu2009} computed as function of the number of clusters in $k$-means, for \mbox{$k \in \{100, 200, \cdots, 1200\}$}. Results on MPEG-7 CE-1 are presented in Fig.~\ref{fig:size_and_bandwidth} (a). The $k$ vs. MAP chart for MPEG-7 CE-2 is equivalent.

The bandwidth $h$ of the mean shift clustering procedure was chosen after assuming \mbox{$h \in \{0.1, 0.3, 0.5, 0.7, 0.9\}$} for MPEG-7 CE-1 and \mbox{$h \in \{1.1, 1.3, 1.5, 1.7, 1.9\}$} for MPEG-7 CE-2 and then analyzing the respective MAP values. A small bandwidth value produces a codebook with an increased number of labels, which makes it more detailed in terms of nuances between visual hierarchies. On the other hand, large $h$ values produce codebooks with fewer labels, which makes them more resilient to small changes. The parameter $h$ was set to $0.7$ and $1.7$ for, respectively, MPEG-7 CE-1 and MPEG-7 CE-2 databases. Both values maximized the MAP value (see Fig.~\ref{fig:size_and_bandwidth} (b) for the former) and produced codebooks having $25$ and $21$ labels, respectively. According to our experience, the advantage of using mean shift with the RBF kernel instead of $k$-means for visual hierarchies codebook learning is that it is quite more difficult to set the expected number of clusters~($k$) for visual words representing trademark images than for simpler shapes.

\stepskip
\noindent
\textbf{CNN-based Implementation and Training.} The CNN-based approach inspired by Perez's~\emph{et al.}~\cite{Perez2018} work consists of a VGG16 network~\cite{Simonyan2014} in which the last fully connected layer was replaced by a $70$ neurons layer and a $51$ neurons layer, each corresponding to the number of classes included in, respectively, the MPEG-7 CE-1 and MPEG-7 CE-2 databases. The networks were implemented using \texttt{TensorFlow}\footnote{TensorFlow: https://www.tensorflow.org/} and \texttt{Keras}\footnote{Keras: https://keras.io/} frameworks.

We have used the same optimization hyperparameters for both networks. They were trained with Stochastic Gradient Descent with $0.9$~of momentum factor, batch size of~$32$, and fixed learning rate of~$0.0001$. They were initialized with weights trained on the \emph{ImageNet Dataset}~\cite{DengEtAl2009} and fine-tuned with the train data of MPEG-7 CE-1 and MPEG-7 CE-2 databases, respectively. We performed data augmentation by randomly rotating, zooming, shifting, and flipping the testing database entries. The networks were trained in two phases. In the first phase, only the replaced layer of the networks was fine-tuned over $50$~epochs. In the second phase, we performed fine-tuning of the whole networks over $20$~epochs.

\stepskip
\noindent
\textbf{Evaluation.} We used $10$-fold cross-validation to assess the performance of the compared solutions on TIR tasks. We distributed the images randomly in the folds while keeping uniform distribution per class in each of them.

Precision values were computed as function of recall for the techniques proposed by ZM~\cite{Khotanzad1990}, Anuar \emph{et al.}~\cite{Anuar2013}, Liu \emph{et al.}~\cite{Liu2017}, and our approach. For those techniques, we computed the $\textrm{F}_1$~score by taking precision at recall of 100\%. The computation of $\textrm{F}_1$~score of the CNN-based approach is straightforward.

\subsection{Precision-Recall Analysis}
\label{sec:precision-recall_analysis}

\begin{figure}[t]
    \centering
    \includegraphics[width=\columnwidth]{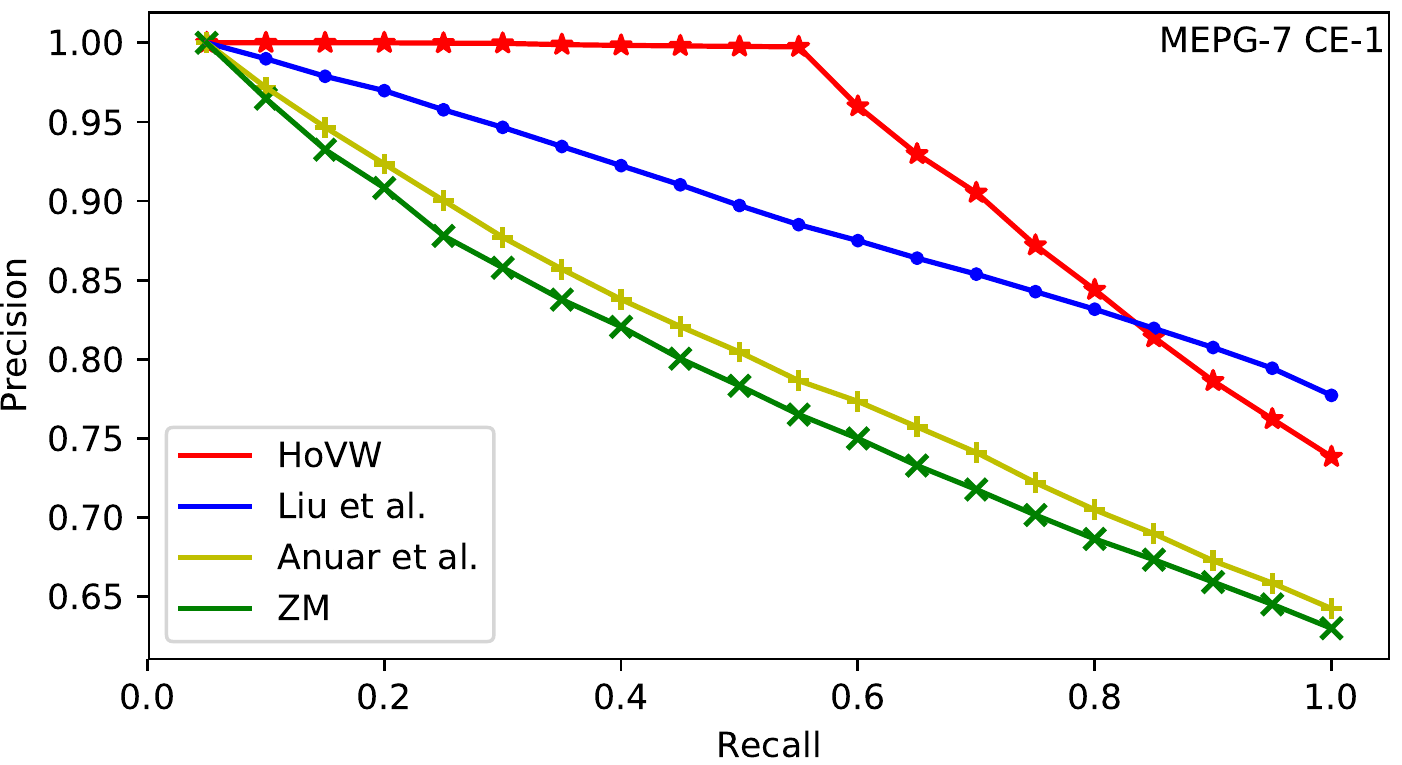}
    \includegraphics[width=\columnwidth]{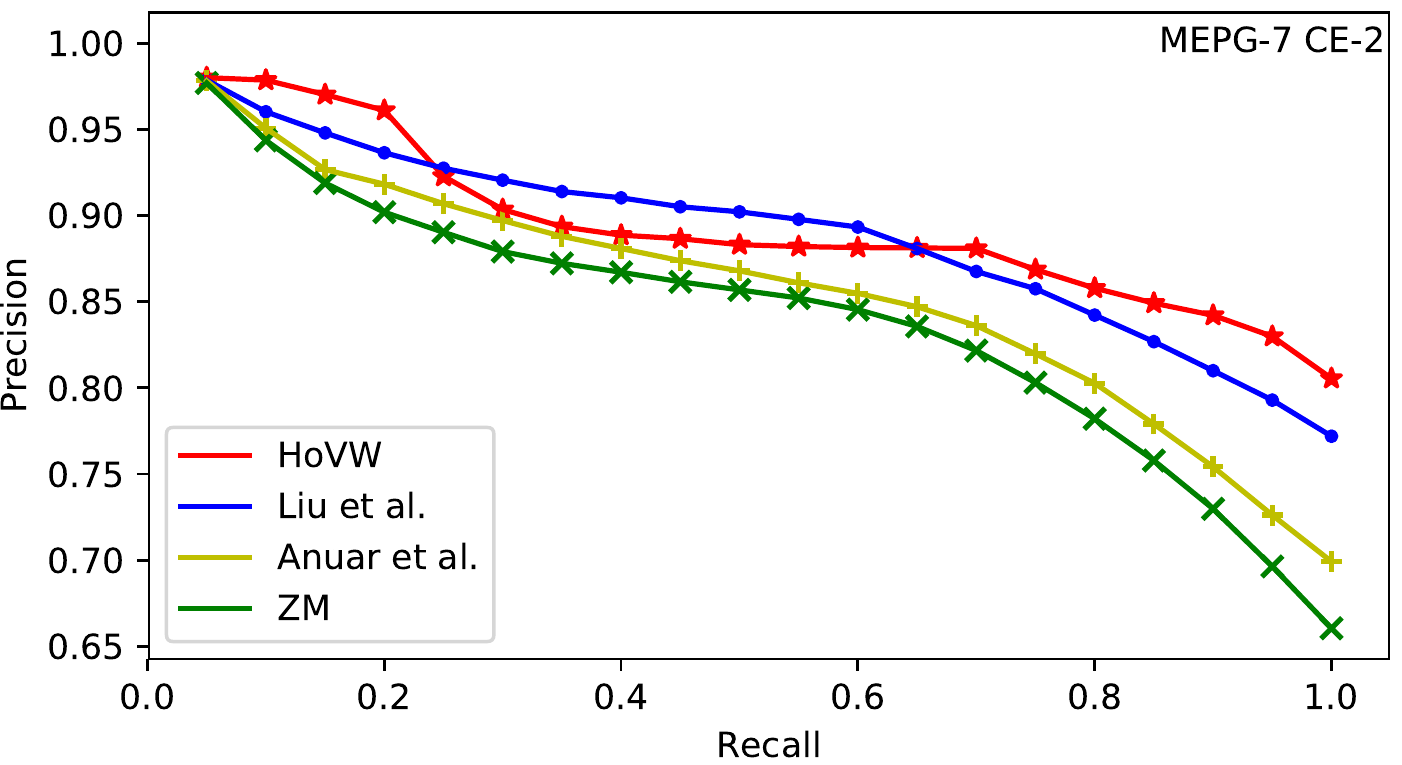}
    \caption{The precision-recall curves of compared approaches.}
    \label{fig:precision-recall}
\end{figure}

In experiments on MPEG-7 CE-1 (Fig.~\ref{fig:precision-recall}, top), HoVW yields a near-perfect result for the first $11$ images retrieved in all categories, obtaining precision of $99.79\%$ up to recall of $55\%$. For the last correlated image retrieved, HoVW's precision was $72\%$. The decay of the curve is easily explained by the way database entries are retrieved by HoVW and how we compute precision. As described in Section~\ref{sec:hovw}, HoVW retrieves from the database all images having the same label as the query hierarchy and, gradually, chunks of entries with close labels. Conservatively, we include in the calculation of precision and recall all entries associated with the secondary sets of retrieved images. Therefore, it is expected that precision will decline with the increase of recall as the labels containing the remaining similar items become further from the original label. For instance, consider the example on Fig.~\ref{fig:similar-search} and a total of three similar images to retrieve, in which the green arrows point to correctly retrieved images, while the red arrow indicates the erroneous retrieved images. For the $\alpha$ cluster, two of the total similar images and one non-similar image were retrieved, accounting a mean precision and recall of~$66\%$. When cluster $\beta$ is reached, the precision steadies on~$66\%$, while the recall reaches~$100\%$. We believe that the adoption of specialized data structures to manage intra-label relationships would help mitigate this issue.

When compared to Anuar~\emph{et al.} and ZM, it can be seen in Fig.~\ref{fig:precision-recall} (top) that the proposed approach has the best precision-recall curve in all ranks. HoVW outperforms Liu \emph{et al.}'s approach up to recall of $80\%$ and their performance are comparable from a recall of $85\%$ to $100\%$. This result suggests that in practical applications of TIR specialists would have access to the expected top-ranked similar images while judging trademark infringement.

Fig.~\ref{fig:precision-recall} (bottom) shows comparable results on the MPEG-7 CE-2 database for the competing techniques. In this database, HoVW presents slightly better performance than Liu \emph{et al.}'s approach on $11$ out $20$ cases of the precision-recall curve and outperforms both Anuar~\emph{et al.} and ZM techniques.

\subsection{$\textrm{F}_1$~Score Analysis}
\label{sec:f1score_analysis}

In this analysis, we used a state-of-the-art CNN-based approach as a baseline for comparison. Table~\ref{tab:f1score} shows the $\textrm{F}_1$~score obtained by each compared technique on each database. On the MEPG-7 CE-1, the CNN-based approach accomplishes better $\textrm{F}_1$~score, reaching 88\%, while HoVW reaches up to 85\%. However, on MEPG-7 CE-2, HoVW outperforms the CNN-based approach by achieving a $\textrm{F}_1$~score of 89\%.

\begin{table}[t]
    \caption{The $\textrm{F}_1$~score obtained by each approach on the MPEG-7 CE-1 and MPEG-7 CE-2 databases.}
    \label{tab:f1score}
    \centering
    \begin{tabular}{ccc}
    \toprule
    \textbf{Approach}     & \textbf{MPEG-7 CE-1} & \textbf{MPEG-7 CE-2} \\
    \midrule
    HoVW & 0.85 & \textbf{0.89}  \\
    Liu~\emph{et al.}~\cite{Liu2017} & 0.87 & 0.87 \\
    Anuar~\emph{et al.}~\cite{Anuar2013} & 0.78 & 0.82 \\
    ZM & 0.77 & 0.79 \\
    CNN-based & \textbf{0.88} & 0.81 \\
    \bottomrule
    \end{tabular}
\end{table}

\subsection{Limitations}
\label{sec:limitations}

Recall that trademark images may include graphical or figurative shape patterns, text words, or both. The HoVW does not include the analysis of text in its conception. Also, the image decomposition does not consider Gestalt. As a result, some implicit organization of shape components may be disregarded while building the hierarchy of component shapes.

\section{Conclusion and Future Works}
\label{sec:conclusion}

We proposed a learning-based approach for TIR that uses two codebooks. The first codebook encodes basic shapes expected in the images using $29$-dimension feature vectors combining region-based and contour-based descriptors. The second codebook encodes both local and global information of trademark images through hierarchical arrangements of their component shapes. The hierarchy is defined as a tree where each node is related to a component shape while tree levels describe the topological relationship of the components. Tree dissimilarity is computed using an efficient tree edit distance algorithm proposed by Pawlik and Augsten~\cite{Pawlik2016}. The main contributions of our work are a new learning-based framework for the hierarchical representation of elements in binary images, and its application on trademark image description and retrieval from image databases. Experimental results on well-known image databases show that our approach outperforms state-of-the-art techniques.

As future work, we are exploring ways to incorporate optical character recognition and principles from Gestalt psychology while decomposing trademark images into basic shapes. The Gestalt properties could be useful to enhance the results obtained from the decomposition, providing normalized shapes grounded to a theory. Also, we are investigating whether state-of-the-art hierarchical image segmentation is suitable for the proposed technique.

\iffinal

\section*{Acknowledgments}
This work was partially supported by the Brazilian Council for Scientific and Technological Development~(CNPq -- Grant \mbox{311.037/2017-8}) and the Rio de Janeiro State Foundation to Support Research~(FAPERJ -- Grant \mbox{E-26/202.718/2018}) agencies. V\'{i}tor N. Louren\c{c}o was sponsored by a CNPq fellowship.

\fi

\bibliographystyle{IEEEbib}
\bibliography{references}

\end{document}